\documentclass[11pt]{article}

\usepackage[preprint]{acl}

\usepackage{times}
\usepackage{latexsym}

\usepackage[T1]{fontenc}

\usepackage[utf8]{inputenc}

\usepackage{microtype}

\usepackage{inconsolata}

\usepackage{graphicx}
\usepackage{booktabs}
\title{Evaluating Accounting Reasoning Capabilities of Large Language Models}

\author{Jie Zhou,  Xin Chen, Jie Zhang, Hai Li, Jie Wang, Zhe Li \\
School of Computer Engineering, Jiangsu Ocean University \\
jzhou23@jou.edu.cn, lizhe@jou.edu.cn
}

\begin{document}
\maketitle
\begin{abstract}
Large language models are transforming learning, cognition, and research across many fields. Effectively integrating them into professional domains, such as accounting, is a key challenge for enterprise digital transformation. To address this, we define vertical domain accounting reasoning and propose evaluation criteria derived from an analysis of the training data characteristics of representative GLM models. These criteria support systematic study of accounting reasoning and provide benchmarks for performance improvement. Using this framework, we evaluate GLM-6B, GLM-130B, GLM-4, and OpenAI GPT-4 on accounting reasoning tasks. Results show that prompt design significantly affects performance, with GPT-4 demonstrating the strongest capability. Despite these gains, current models remain insufficient for real-world enterprise accounting, indicating the need for further optimization to unlock their full practical value.
\end{abstract}

\section{Introduction}

With the rapid progress of artificial intelligence, including big data, supercomputing, brain inspired intelligence, and large language models, general purpose language models have drawn strong interest from both academia and industry. Artificial intelligence has become a central driver of technological innovation, industrial upgrading, and national competitiveness. In China, policies such as the New Generation Artificial Intelligence Development Plan and the Interim Measures for the Administration of Generative Artificial Intelligence Services have emphasized interdisciplinary research, responsible governance, and large scale application, creating favorable conditions for integrating large language models into professional domains such as accounting.

As large language models mature, their development focus has shifted from model scale to deep application integration. Effectively embedding these models into professional workflows and unlocking domain specific value has become a key research challenge. Compared with narrowly specialized systems, general purpose language models show advantages in accounting tasks through broader knowledge coverage, stronger prompt comprehension, and flexible reasoning. These capabilities support applications such as financial analysis, taxation, auditing, and management support. However, systematic domain specific accounting models are still underdeveloped, and current practical performance remains limited.

Most existing accounting applications rely on prompt engineering to improve model outputs. While this approach can yield short term gains, it provides limited insight into internal reasoning processes and offers little support for systematic optimization. Accounting tasks place high demands on mathematical reasoning, logical consistency, and symbolic computation, which are not fully addressed by prompt based methods alone.
This study argues that reasoning ability is the core factor underlying large language model performance in accounting. We therefore propose a structured analysis and evaluation of accounting reasoning capabilities. By integrating accounting expertise with reasoning oriented evaluation tasks, we construct a dedicated benchmark and analyze experimental results to identify opportunities for improvement.

Our contributions are four-fold. We introduce the concept of vertical domain accounting reasoning and clarify its relationship with mathematical and logical reasoning. We develop a comprehensive benchmark covering accounting principles, financial reporting, cost accounting, and auditing. We evaluate representative models, including GPT 4 and GLM series models, and analyze their strengths and limitations. Finally, we explore strategies that combine domain knowledge with reasoning optimization to advance intelligent accounting systems and support real world deployment.

\section{Related Work}

\subsection{Development of LLMs in the Accounting Domain}

The effective use of general purpose large language models in vertical fields such as medicine and law offers useful insights for their adoption in accounting. In medical settings, LLMs have been applied to tasks including clinical decision support, medical image interpretation, and report generation \citep{panagiotou2024}. Owing to the strong interpretability of medical images, these models can capture visual patterns and, when combined with natural language generation, enhance system transparency and explainability, which supports improved clinical decision making \citep{chalkidis2023}.

In contrast, applications in the legal domain are typically limited to assistive roles, as legal systems are complex, rule based, and highly regulated. Automated reasoning must conform to legal and ethical constraints, which restricts the extent of model autonomy. Accounting shares a similar rule driven structure but places heavier demands on numerical precision, quantitative reasoning, and cross statement consistency. Unlike vision centered medical tasks, accounting requires models to retain formal standards, track causal relationships, and perform multi step logical and numerical reasoning. Consequently, robust reasoning ability is especially important for accounting use cases.

Existing work can be grouped into several lines of research. One line seeks to strengthen general reasoning skills, such as mathematical or logical reasoning \citep{huang2022reasoning}. These methods often treat reasoning as independent of domain knowledge, which limits their effectiveness in accounting tasks that rely heavily on professional expertise \citep{li2023}. Dedicated studies on accounting reasoning remain scarce.
Another line focuses on domain adaptation through specialized data or larger models. Examples include BloombergGPT \citep{wu2023bloomberggpt} and FinGPT \citep{liu2023fingpt}, which leverage financial corpora or increased parameter scales. While these models can improve task performance, they do not consistently enhance reasoning ability and often lack interpretability, making it unclear whether gains reflect true reasoning or shallow pattern learning \citep{zhao2023}.
A third line emphasizes evaluation based on final answer accuracy, such as financial question answering or knowledge retrieval \citep{faturos2023,theuma2024,shah2023}. These text oriented tasks overlook the structured and sequential nature of accounting reasoning and therefore fail to measure core reasoning competence.
Finally, several exploratory studies examine the potential role of LLMs in accounting practice \citep{he2023,ouyang2024}. Although informative, these efforts have yet to define systematic evaluation frameworks or concrete optimization methods for accounting reasoning.

In summary, most existing research applies LLMs to surface level language tasks in accounting. Comprehensive benchmarks and evaluation frameworks that directly assess accounting reasoning capabilities are still lacking.

\subsection{Reasoning Capability of LLMs in Accounting}

In LLM research, reasoning is generally understood as the capacity to derive conclusions from provided facts and constraints \citep{huang2022reasoning,yu2023}. Most reasoning evaluations focus on output correctness in inference-based tasks. From a structural perspective, reasoning is commonly divided into mathematical, logical, causal, and commonsense reasoning \citep{sun2023,huang2022reasoning}. Among these, mathematical and logical reasoning are the most relevant to accounting applications.

Accounting tasks frequently require precise numerical computation, multi-step deduction, and adherence to well-defined rules. As a result, reasoning ability is a key determinant of LLM effectiveness in this domain. Logical reasoning assesses whether a model can draw valid conclusions from premises under explicit rules, often evaluated using natural language inference benchmarks \citep{srivastava2022,yu2023,chang2023}. However, such benchmarks are designed for general-purpose language understanding and fail to capture the structured, rule-driven nature of accounting reasoning.

Mathematical reasoning measures a model’s ability to carry out arithmetic and symbolic operations. Benchmarks such as GSM8K \citep{cobbe2021} evaluate this capability through multi-step math word problems that require intermediate value tracking, closely mirroring accounting-style calculations.

In practice, most accounting tasks rely on chained numerical operations rather than high-level financial interpretation. Errors in intermediate result propagation can therefore compound and lead to incorrect final outcomes, making robust multi-step mathematical reasoning especially important. In addition, many accounting problems involve time-dependent entities and constraints. Incorporating temporal knowledge graphs \citep{xiongtilp,xiong2024teilp,xiong2024large} into LLM-based symbolic reasoning frameworks \citep{yang2024harnessing} enables structured state maintenance and improves consistency across reasoning steps.

Recent work has explored structure-aware approaches to strengthen multi-step reasoning. SWAP \citep{xiong2025deliberate} introduces planning mechanisms that explicitly model reasoning structure. MR-GSM8K \citep{zeng2024} further increases reasoning complexity by filtering out short-solution problems and emphasizing longer computation chains, yielding tasks that more closely resemble real accounting scenarios. After additional filtering, 586 problems requiring at least three computational steps remain. This subset provides a suitable basis for evaluating multi-step numerical reasoning in accounting settings and is referred to as the Multi-Calculation Benchmark.

\subsection{Evaluation Benchmarks for Chinese Accounting Reasoning}

Benchmarks for evaluating Chinese accounting reasoning mainly come from two directions. One line of work adapts general Chinese natural language inference benchmarks translated from English, while the other relies on datasets drawn from accounting-related examinations and professional qualification tests. Compared with multi-step mathematical reasoning, evaluating accounting reasoning in Chinese presents additional difficulties due to variations in accounting standards, linguistic structure, and task design.

The Chinese Language Understanding Evaluation (CLUE) benchmark \citep{xu2020}, often viewed as the Chinese analogue of GLUE \citep{wang2018}, provides a broad assessment of Chinese language understanding through tasks such as classification, reading comprehension, and natural language inference. Despite its coverage, CLUE focuses on general linguistic capability and does not explicitly target accounting reasoning.

Within CLUE, CMNLI evaluates entailment, contradiction, and neutrality between a premise and a hypothesis. OCNLI, developed natively in Chinese, better captures Chinese language usage and reasoning characteristics. ACMC further adapts CMNLI and OCNLI to financial and accounting-related scenarios, but it still lacks structured numerical computation and explicit accounting reasoning chains.

To better reflect real accounting practice, some studies turn to professional examination questions, including those from the Chinese Certified Public Accountant (CPA) exams. These questions span accounting treatments, financial calculations, auditing judgments, and tax-related reasoning. However, CPA problems often involve long descriptions and substantial domain knowledge, which makes evaluation difficult. In addition, model performance is highly dependent on prompt formulation, and errors may arise from misinterpretation rather than weak reasoning ability.

Overall, existing benchmarks either emphasize general reasoning without accounting specificity or include accounting content without a systematic evaluation of reasoning processes. This gap motivates the need for accounting-focused reasoning benchmarks that combine domain knowledge with structured numerical inference. This work responds to this need by constructing dedicated accounting evaluation datasets and analyzing LLM performance across multiple reasoning dimensions.

\section{Experiments}

\subsection{Experimental Setup}

This work examines the accounting reasoning abilities of large language models. We define accounting reasoning as the joint use of logical inference and numerical computation within accounting-specific rules.
Consistent with prior studies, accounting reasoning is viewed as a combined capability involving arithmetic computation, rule application, and coherent multi-step deduction. Accordingly, model performance is assessed from multiple angles, including stepwise calculation accuracy, understanding of accounting principles, and overall reasoning quality in realistic problem settings.

Empirical findings from the Stanford Alpaca project show that carefully curated datasets, even when reduced to about 5 percent of the original size, can substantially improve model performance \citep{xia2024}. Data quality therefore plays a key role in strengthening reasoning accuracy and stability. Following this insight, we first evaluate baseline reasoning ability using general-purpose benchmarks, and then conduct targeted assessments on accounting-specific datasets.

To provide a structured analysis, the evaluation is organized into three levels: mathematical reasoning, accounting knowledge reasoning, and integrated accounting reasoning. These levels reflect increasing task complexity and practical relevance, progressing from basic numerical operations to scenario-driven accounting analysis.

\begin{table}[t]
\centering
\small
\caption{Benchmark Datasets and Reasoning Tasks}
\label{tab:benchmark_tasks}
\begin{tabular}{p{0.3\linewidth} p{0.58\linewidth}}
\toprule
\textbf{Reasoning Task} & \textbf{Datasets} \\
\midrule
Mathematical and Computational Reasoning 
& GSM8K, SVAMP, ASDiv, AQuA-RAT, MAWPS, AddSub, MultiArith, SingleEq, SingleOp \\
\midrule
Logical Reasoning 
& ProofWriter, EntailmentBank, RuleTaker, CLUTRR, FLD \\
\midrule
Commonsense Reasoning 
& CommonsenseQA, StrategyQA, ARC, SayCan, BoolQ, HotpotQA, OpenBookQA, PIQA, WikiWhy, COPA \\
\midrule
Abstract Reasoning 
& Last Letter Concatenation, Coin Flip \\
\midrule
Other 
& SNLI, MultiNLI, HellaSwag, SQuAD, BIG-bench, SCAN, BBH \\
\bottomrule
\end{tabular}
\end{table}

\subsection{Evaluation of Mathematical Reasoning Ability}

A range of benchmarks has been proposed to assess the mathematical reasoning capabilities of LLMs, including datasets for natural language inference and multi-step arithmetic reasoning \citep{srivastava2022,yu2023,chang2023}. Despite their popularity, most general benchmarks do not reflect the computational characteristics of accounting tasks, particularly those involving conditional rules and the accumulation of intermediate errors.

GSM8K \citep{cobbe2021} is one of the most commonly adopted datasets for evaluating multi-step numerical reasoning. 
Nevertheless, many GSM8K problems involve relatively short reasoning paths. To emphasize longer reasoning chains, MR-GSM8K \citep{zeng2024} is derived by removing problems with brief solution trajectories and retaining those that require extended computation. Specifically, instances with answers shorter than 300 words are excluded, yielding a dataset that stresses sustained reasoning and cumulative error effects.

After additional filtering, 586 MR-GSM8K instances that require at least three computation steps are retained. This subset serves as our multi-step calculation evaluation benchmark, referred to as the \textit{Multi-Calculation Benchmark}, and is well suited for examining the robustness of LLMs in accounting-style numerical reasoning.

\subsection{Evaluation of Chinese Accounting Reasoning Ability}

Assessing accounting reasoning in Chinese involves additional complexity arising from linguistic structure, accounting regulations, and specialized terminology. Current Chinese reasoning benchmarks mainly fall into two categories: translated general-purpose reasoning datasets and datasets drawn from professional accounting examinations.

The Chinese Language Understanding Evaluation (CLUE) benchmark \citep{xu2020} is commonly used to measure Chinese language comprehension and reasoning. It includes tasks such as classification, reading comprehension, and natural language inference, and is often viewed as the Chinese equivalent of GLUE \citep{wang2018}. While CLUE provides broad coverage of reasoning skills, it is not designed to evaluate accounting-specific inference.

To better reflect accounting practice, we also consider datasets derived from professional qualification tests, including the Chinese Certified Public Accountant (CPA) examinations. CPA questions span accounting treatments, numerical computation, auditing decisions, and taxation, requiring models to combine arithmetic reasoning with regulatory knowledge.
Despite their relevance, CPA-based evaluations pose difficulties. The questions often involve lengthy contexts, dense technical language, and implicit assumptions, which can cause errors stemming from misinterpretation rather than weak reasoning. In addition, performance is sensitive to prompt formulation, complicating the isolation of pure reasoning ability.

\subsection{Comprehensive Accounting Reasoning Evaluation}

To obtain a more robust evaluation, this work integrates general reasoning benchmarks, accounting-oriented computation tasks, and questions modeled after professional examinations. Examining performance across these settings allows us to better characterize the capabilities and limitations of LLMs in accounting reasoning.

The results show that LLMs perform well on basic calculations and shallow reasoning tasks, but accuracy drops noticeably as reasoning chains become longer. Errors introduced at intermediate steps tend to compound, and maintaining correct application of accounting rules becomes difficult in scenarios involving multiple interacting constraints.

In summary, current LLMs exhibit encouraging baseline reasoning skills, yet they remain insufficient for practical accounting use cases. Closing this gap will require improvements beyond prompt engineering, including higher-quality domain data and model designs that explicitly support structured and domain-aware reasoning.

\begin{figure}[t]
  \centering
  \includegraphics[width=0.9\linewidth]{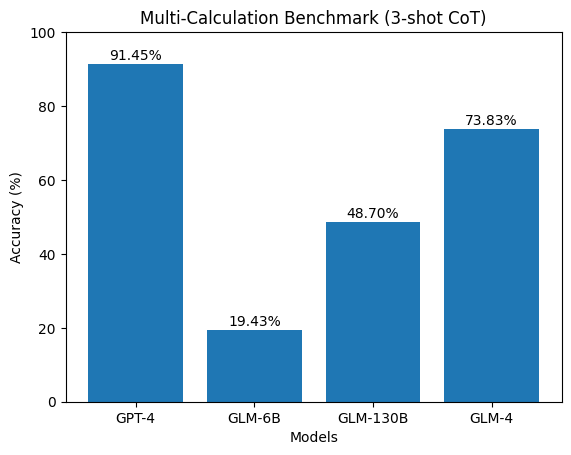}
  \caption{Multi-Calculation Benchmark: 3-shot Chain-of-Thought Evaluation Results.}
  \label{fig:2}
\end{figure}

\subsection{Evaluation Results and Analysis}

We evaluate the performance of OpenAI GPT-4 and compare it with GLM-6B, GLM-130B, and GLM-4. Chain-of-Thought (CoT) prompting is applied during evaluation to encourage explicit generation of intermediate reasoning steps, which improves both reasoning quality and interpretability. In addition, Few-shot Learning is used by providing representative example solutions to guide the reasoning and answer generation process.

We first examine model behavior under different prompting strategies, including Zero-shot, Few-shot, CoT, and Zero-shot-CoT. The results indicate that Zero-shot and Zero-shot-CoT outputs frequently exhibit logical inconsistencies. In many cases, models generate a plausible final answer while the intermediate reasoning does not fully support the conclusion. This suggests that evaluation based solely on final answers can overestimate reasoning quality. To address this issue, we adopt Few-shot-CoT prompting for the main evaluation. Compared with Zero-shot settings, Few-shot prompting leads to substantial accuracy gains, especially on tasks requiring multi-step reasoning. Empirically, Few-shot-CoT improves accuracy by roughly 50 percent. In accounting-related tasks, this strategy enables models to correctly handle problems involving up to nine arithmetic operations, reflecting a marked improvement in reasoning depth.

\begin{figure}[t]
  \centering
  \includegraphics[width=0.9\linewidth]{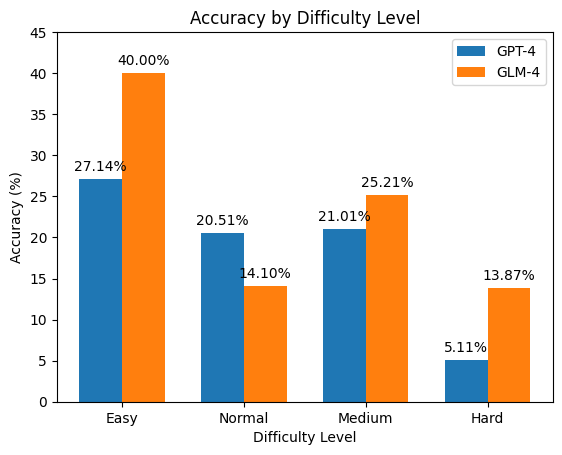}
  \caption{Accounting-Reasoning-Benchmark (3-shot CoT) Evaluation Results.}
  \label{fig:3}
\end{figure}

\begin{figure*}[t]
  \centering
  \includegraphics[width=0.9\linewidth]{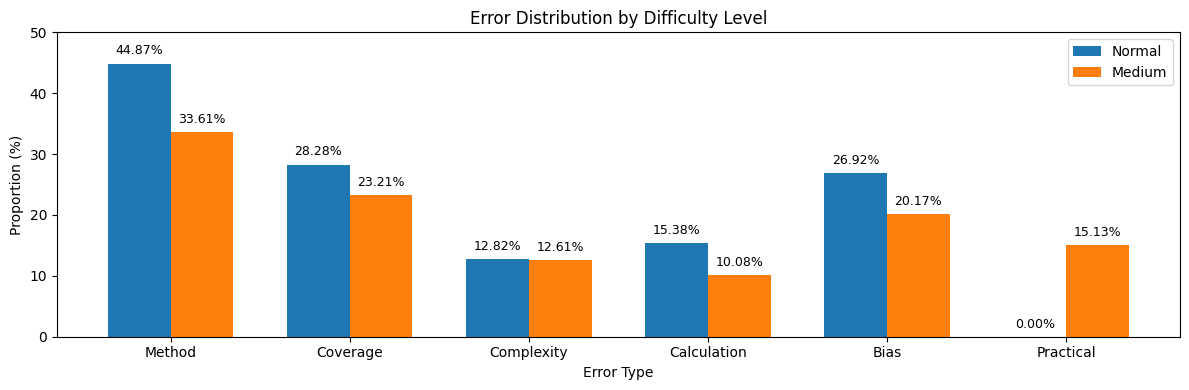}
  \caption{Comparison of Error Type Proportions between Normal and Medium Difficulty Levels.}
  \label{fig:4}
\end{figure*}

Based on these observations, Few-shot-CoT is adopted as the default evaluation setting. We follow a GSM8K-style Few-shot-CoT format that explicitly separates step-by-step reasoning from the final answer. For instance, when solving accounting problems involving asset acquisition and depreciation, the model first presents a structured reasoning process and then reports the final numerical result. This format improves interpretability and ensures consistent evaluation.

After collecting model outputs, quantitative evaluation is performed by comparing predictions with ground truth answers. GLM-4 serves as the automatic evaluator for responses generated by GLM-6B, GLM-130B, and GPT-4. Overall accuracy is computed across all questions and reported for the Multi-Calculation Benchmark.

GPT-4 achieves the highest overall accuracy among the evaluated models, establishing a strong reference point for accounting reasoning. Nevertheless, GPT-4 still fails on a subset of complex accounting problems, indicating room for improvement. Smaller models face greater difficulty with multi-step arithmetic reasoning. GLM-6B attains an accuracy of roughly 20 percent, while GLM-130B reaches about 60 percent, both falling short of the accuracy levels typically required in professional accounting practice. Although GLM-4 performs competitively, its accuracy remains below the desired threshold, underscoring the limitations of current LLMs in precise numerical reasoning.

Since the benchmark emphasizes multi-step arithmetic accounting problems, only GPT-4 and GLM-4 are further evaluated on the Accounting Reasoning Benchmark. The results show that GPT-4 achieves an accuracy of 16.58 percent, whereas GLM-4 attains 21.78 percent, suggesting a slight advantage for GLM-4 in accounting knowledge reasoning. As the number of computation steps increases, performance deteriorates for both models, and neither demonstrates consistently reliable accounting reasoning.

Error analysis reveals several recurring failure patterns. One major category involves incorrect application of accounting logic, such as misinterpreting asset treatments, standards, or tax rules. Another common issue is incomplete domain knowledge, where critical conditions or constraints are overlooked. Models also struggle with problems requiring intertwined accounting procedures, failing to track dependencies across multiple steps. Arithmetic and logical inconsistencies arise when reasoning structures are sound but numerical calculations are incorrect. Additional errors stem from vague conceptual understanding or misinterpretation of foundational accounting principles.

A finer-grained analysis shows that misunderstandings of accounting principles and insufficient conceptual coverage together account for more than half of all errors. Mistakes related to detailed procedures, such as bookkeeping or financial statement preparation, occur less frequently but remain non-negligible. Principle-level errors are particularly harmful because they often propagate throughout the reasoning process, whereas procedural errors tend to affect only localized steps. This pattern mirrors the emphasis on conceptual mastery in professional accounting examinations.

Overall, the results indicate that current LLMs still lack robust accounting reasoning capabilities. While they offer useful assistance for accounting-related tasks, substantial progress is needed in domain knowledge integration, reasoning reliability, and numerical precision before deployment in professional accounting environments becomes viable.

\section{Conclusion}

This study examines accounting-related reasoning in large language models using two dedicated benchmarks. The results indicate that although LLMs handle general multi-step reasoning effectively, performance degrades markedly on accounting-focused tasks, revealing a substantial gap between broad reasoning skills and domain-specific competence. Further error analysis shows persistent difficulties in correctly applying accounting principles in complex scenarios. Overall, these findings suggest that current LLMs remain unsuitable for dependable use in professional accounting settings.

\section*{Limitations}

This work has several constraints. The benchmark emphasizes structured accounting reasoning and therefore does not reflect the full range of real accounting practice. In addition, the evaluation covers only a small set of representative LLMs, so the findings may not extend to models with different architectures or specialized training data. Finally, the analysis relies on prompt-based reasoning, making results sensitive to prompt formulation. Future research should address these issues by expanding benchmark coverage, evaluating a wider variety of models, and exploring systematic approaches to domain adaptation.

\bibliography{custom}

\end{document}